\documentclass{article}
\usepackage{ijcai25}

\usepackage{times}
\usepackage{soul}
\usepackage{url}
\usepackage[hidelinks]{hyperref}
\usepackage[utf8]{inputenc}
\usepackage[small]{caption}
\usepackage{graphicx}
\usepackage{amsmath}
\usepackage{amsthm}
\usepackage{booktabs}
\usepackage{algorithm}
\usepackage{algorithmic}
\usepackage[switch]{lineno}
\usepackage[table]{xcolor}
\usepackage{amssymb}
\usepackage{subcaption}

\title{Enhancing Chemical Reaction and Retrosynthesis Prediction with Large Language Model and Dual-task Learning}

\author{
Xuan Lin$^{1}$\and
Qingrui Liu$^1$\and
Hongxin Xiang$^2\textsuperscript{*}$\and
Daojian Zeng$^3$\And
Xiangxiang Zeng$^2$\\
\affiliations
$^1$School of Computer Science, Xiangtan University\\
$^2$College of Computer Science and Electronic Engineering, Hunan University\\
$^3$Institute of AI and Targeted International Communication, Hunan Normal University\\
\emails
jack\_lin@xtu.edu.cn,
xianghx@hnu.edu.cn
}
\begin{document}

\maketitle
\let\thefootnote\relax
\footnotetext{
  \textsuperscript{*}\parbox[t]{\textwidth}{
    \textcolor{black}{means the corresponding author.
    % //
    % The full version is at \url{https://arxiv.org/} (on hold).
    }
  }
}

% \maketitle

\begin{abstract}

\textcolor{black}{Chemical reaction and retrosynthesis prediction are fundamental tasks in drug discovery. Recently, large language models (LLMs) have shown potential in many domains. However, directly applying LLMs to these tasks faces two major challenges: (i) lacking a large-scale chemical synthesis-related instruction dataset; (ii) ignoring the close correlation between reaction and retrosynthesis prediction for the existing fine-tuning strategies.
To address these challenges, we propose ChemDual, a novel LLM framework for accurate chemical synthesis. Specifically, considering the high cost of data acquisition for reaction and retrosynthesis, ChemDual regards the reaction-and-retrosynthesis of molecules as a related recombination-and-fragmentation process and constructs a large-scale of 4.4 million instruction dataset.
Furthermore, ChemDual introduces an enhanced LLaMA, equipped with a multi-scale tokenizer and dual-task learning strategy, to jointly optimize the process of recombination and fragmentation as well as the tasks between reaction and retrosynthesis prediction. 
Extensive experiments on Mol-Instruction and USPTO-50K datasets demonstrate that ChemDual achieves state-of-the-art performance in both predictions of reaction and retrosynthesis, outperforming the existing conventional single-task approaches and the general open-source LLMs. Through molecular docking analysis, ChemDual generates compounds with diverse and strong protein binding affinity, further highlighting its strong potential in drug design.}
\end{abstract}

\section{Introduction}

\begin{figure*}
    \centering
    \includegraphics[width=0.9\linewidth]{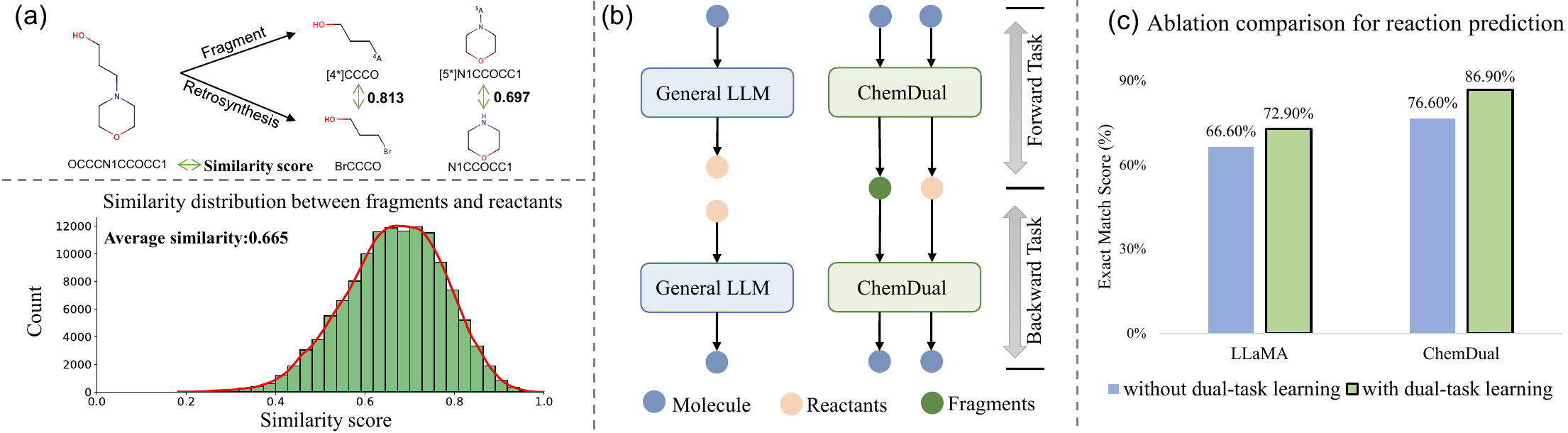}
    \caption{\textcolor{black}{(a) % The lower and upper sub-figures are similarity of "OCCCN1CCOCC1" and similarity distribution of all molecules between fragments and reactants, respectively.
    The upper subgraph is an example of the similarity between fragments and reactants for a molecule "OCCCN1CCOCC1". The lower subgraph is the similarity distribution between fragments and reactants for all molecules. (b) Single-task learning paradigm (left sub-figure) and dual-task learning paradigm of ChemDual (right sub-figure). (c)  The Exact Match score (\%) on the chemical reaction prediction task by using or not using the dual-task learning.}}
    \label{fig:ablation}
\end{figure*}

\textcolor{black}{Chemical reaction and retrosynthesis prediction are fundamental tasks in many fields of chemistry, forming the backbone of synthetic route design, compound optimization, and drug discovery \cite{ucak2022retrosynthetic}. These tasks involve predicting the outcomes of chemical reactions and identifying potential synthetic pathways to create target molecules. Traditional approaches have relied heavily on expert knowledge and manual analysis \cite{liang2024simple}, which are time-consuming and resource-limiting. With the advancement of computational technologies, automated methods have emerged as powerful alternatives to offer significant improvements in speed and efficiency for these critical tasks.}

%\textcolor{blue}{Reaction and retrosynthesis prediction} are fundamental tasks \textcolor{blue}{in many fields of chemistry, especially} drug discovery, forming the backbone of synthetic route design, compound optimization, and the overall efficiency of new drug development. These tasks involve predicting the outcomes of chemical reactions and identifying potential synthetic pathways to create target molecules. Traditional approaches for tackling these problems have relied heavily on expert knowledge and manual analysis, which are not only time-consuming but also limited by human expertise and scalability. With the rapid advancement of computational technologies, automated methods have emerged as powerful alternatives, offering significant improvements in speed and efficiency for these critical tasks.

 \textcolor{black}{Large language models (LLMs) have gained more attention in various domains \cite{chatgpt}, they leverages advanced natural language processing techniques to process and analyze complex biochemical data \cite{chemdfm}. Recently, Mol-Instruction \cite{mol-instruction} has show promising results in addressing the inherent challenges of chemical reaction and retrosynthesis prediction. Despite these advancements, LLMs still face the following two major challenges, which limit their accuracy in predictions of reaction and retrosynthesis compared with traditional machine learning models \cite{ChemLLMBench}.
}

% Despite these advancements, however, LLMs still struggle to accurately interpret specialized chemical knowledge, particularly in the context of simplified molecular input line entry system (SMILES) \cite{weininger1988smiles} notation and molecular structures. As highlighted in studies such as ChemLLMBench \cite{ChemLLMBench}, current LLMs often perform less competitively compared to traditional machine learning models in specific tasks like reaction prediction and retrosynthesis.

\textcolor{black}{\textit{Challenge 1: Lacking a large-scale chemical synthesis-related instruction dataset.} The success of LLM depends on the availability of large and high-quality data \cite{10.5555/3454287.3455083}. In predictions of reaction and retrosynthesis, high-quality data comes from actual chemical synthesis experiments in the laboratory, so it is costly and inefficient, resulting in limited data size. Therefore, this motivates us to find an alternative with low acquisition cost (capable of quickly generating large-scale data) and high data quality (high relevance to real chemical synthesis). As shown in the top of Figure 1(a), we find that fragments generated by breaking of retrosynthetically interesting chemical substructures (BRICS) \cite{BRICS} are highly correlated with reactants. % For example, for the molecule "OCCCN1CCOCC1", we use BRICS to obtain 2 fragments ("[4*]CCCO", "[5*]N1CCOCC1"), which have high similarities (81.3\%, 69.7\%) with the reactants ("BrCCCO" and "N1CCOCC1") from retrosynthesis, respectively. 
We also show the similarity distribution of all molecules in the bottom of Figure 1(a), which has a global average similarity of 66.5\%, indicating that we can use BRICS to generate a large amount of data to assist in the learning of predictions of reaction and retrosynthesis. Therefore, we construct a large-scale chemical synthesis-related instruction dataset, which includes 4.4 million molecules and the corresponding fragments extracted by BRICS.}

\textcolor{black}{\textit{Challenge 2: Ignoring the correlation between predictions of reaction and retrosynthesis.} As shown in the left sub-figure of Figure 1(b), existing methods treat Molecule-to-Reactants and Reactants-to-Molecules as two independent tasks for learning \cite{wang2023retrosynthesis} \cite{liang2023knowledge}. % For example, in the reaction prediction task, the general LLM takes molecules as input to predict the corresponding reactants and is trained by minimizing the gap between the predicted reactants and the true reactants \cite{zhang2021data}. 
However, Molecule-to-Reactants and Reactants-to-Molecules are inverse processes of each other, ignoring the correlation between them will limit LLM in understanding chemical synthesis processes \cite{NIPS2016_dual-learning}. Therefore, we regard Molecule-to-Reactants and Reactants-to-Molecules as forward task and backward task respectively and use a dual-task learning strategy to optimize these two tasks simultaneously. As shown in Figure 1(c), the LLaMA equipped with dual-task learning achieves 6.3\% performance improvement of Exact Match score on reaction prediction task. 
}

\textcolor{black}{
Inspired by these, we propose ChemDual, an enhanced LLaMA equipped with a multi-scale tokenizer and dual-task learning strategy. The multi-scale tokenizer is used to enhance the ability of the model in capturing molecular structures at different scales (such as dummy atom, functional group and fragment, etc.). In dual-task learning, as shown in the right sub-figure of Figure 1(b), we pretrain on molecule-to-fragments and fragments-to-molecule dual tasks constructed from 4.4 million data to enhance ChemDual in understanding of general chemical synthesis, and further fine-tune ChemDual to perform task-specific predictions using molecule-to-reactants and reactants-to-molecule dual tasks. In Figure 1(c), we find that ChemDual achieves better performance compared to LLaMA.} The main contributions of this work are as follows:

%  First, we construct a large-scale dual-task dataset that enables the learning of bidirectional relationships in chemical transformations through molecular fragmentation and recombination. Second, we introduce a dual-task learning strategy that captures the dependencies between forward and backward processes, achieving significant improvements in prediction accuracy for both directions. Third, we conduct extensive experiments to demonstrate that ChemDual outperforms existing models on the Mol-Instruction dataset, particularly in reaction prediction and retrosynthesis tasks. Finally, case studies show the potential of ChemDual in drug discovery, as it generates compounds with favorable binding characteristics and specific molecular interactions.

\begin{itemize}
    \item \textcolor{black}{We construct a large-scale database, which includes 4.4 million molecules and the corresponding fragments, for learning general chemical synthesis-related knowledge.} %, enabling the learning of bidirectional relationships in chemical transformations through molecular fragmentation and recombination.
    \item \textcolor{black}{We propose an enhanced LLM framework based on LLaMA, called ChemDual, equipped with a multi-scale tokenizer and a dual-task learning strategy for learning informative molecular representation from different scales of structures and capturing the correlation between forward and backward processes, respectively.} % achieving significant improvements in prediction accuracy for both directions.
    % \item \textcolor{red}{We propose a novel tokenizer, which helps the ChemDual capture multi-scale structural information.}
    \item \textcolor{black}{Extensive experiments on Mol-Instruction and USPTO-50K show that ChemDual outperforms existing models in predictions of reaction and retrosynthesis. In addition, Case studies demonstrate that ChemDual can generate compounds with favorable binding properties and specific molecular interactions.} % the Mol-Instruction dataset, particularly in reaction prediction and retrosynthesis tasks.
    %, showing the potential of ChemDual in drug discovery. % Case studies show the potential of ChemDual in drug discovery, as it generates compounds with favorable binding characteristics and specific molecular interactions.
\end{itemize}

\begin{figure*}[htbp]
\begin{center}
    \includegraphics[width=\textwidth]{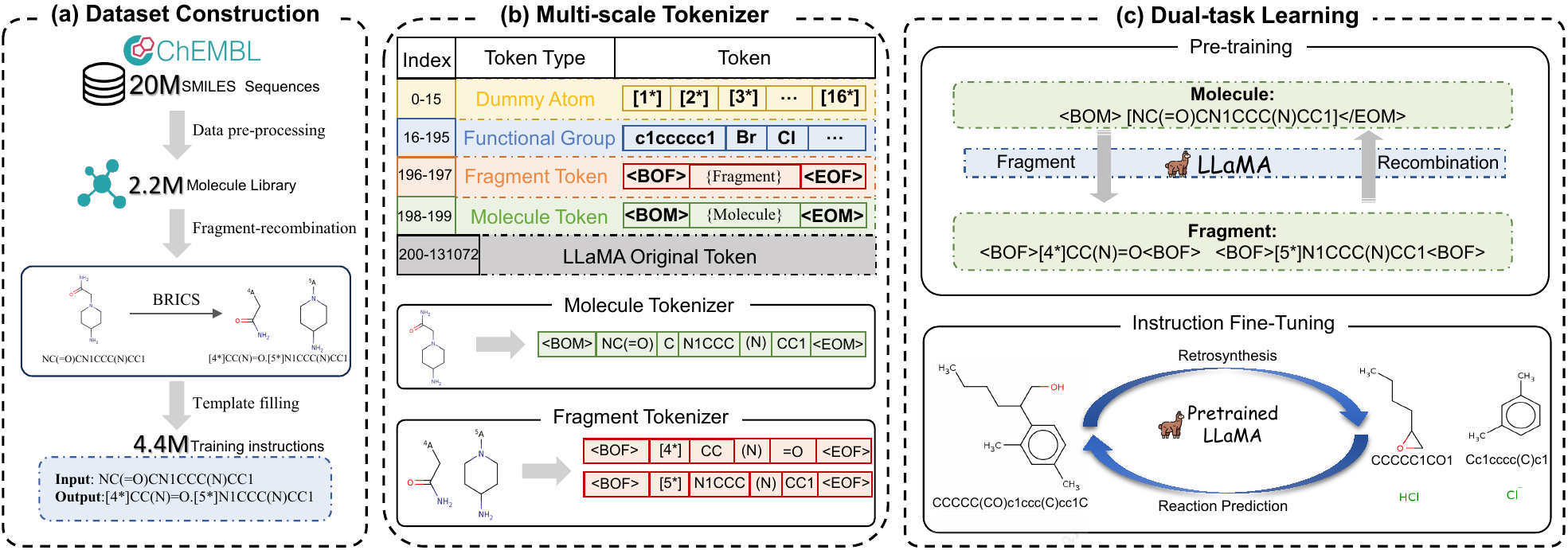}
    %\caption{dual-task Learning For Fine-tuning ChemDual}
    \caption{\textcolor{black}{The overall of proposed ChemDual.}}
    \label{fig:ChemDual}
\end{center}
\end{figure*}
\section{Related Work}

% \subsection{Chemical Large Language Model}
% \subsection{Chemical LLMs with Instruction Fine-tuning}
\textcolor{black}{
Recent researches have proposed a large number of large language models (LLMs), such as LLaMA \cite{touvron2023llamaopenefficientfoundation} and ChatGLM \cite{glm2024chatglm}. Given the important role of chemical reaction and retrosynthesis prediction tasks in drug discovery, many chemical-based domain-specific models have been further proposed \cite{galactica2022,pei2023biot5} The core idea of these models is to use reaction- or retrosynthesis-based instruction datasets to further fine-tune the existing LLMs to enable them to have predictive capabilities for reaction or retrosynthesis. 
% A 使用了什么数据和策略来微调、B 使用了什么数据和策略来微调、C、D ... 
% BioT5+使用的是SELFIES-IUPAC命名转换做预训练提升化学结构理解能力，Mol-Instruction数据集做微调。
BioT5+ \cite{pei2024biot5+} involves molecule a dataset of 30k SELFIES \cite{selfies} mapped to IUPAC names ,with multi-task instruction-based fine-tuning. 
% Text+Chem T5与Mol-Instruction都是引入多分子相关任务(包含反应预测与逆合成)进行联合训练(multitask fine-tuning)
Text+Chem T5 \cite{text+chemt5} and Mol-Instruction \cite{mol-instruction} specifically address chemical tasks by integrating cross-domain knowledge into their frameworks, utilizing a dataset of about 30k samples.
%在传统的深度学习领域一般只能通过增强现有数据集（随机交换反应化合物、生成同一分子不同SMILES序列）来单任务微调。
Retroformer \cite{wan2022retroformer}, trained on the augmentative USPTO-50k \cite{uspto} dataset containing more then 50k reaction samples with a single-task learning strategy, has inspired other works \cite{molecule-transformer} \cite{seo2021gta} that improve performance by incorporating a variety of reasonable SMILES data enhancement methods for the same molecule under a single-task learning framework.
Different from previous methods, % we fully consider the expensiveness and inefficiency of chemical reaction and retrosynthesis data acquisition, 
we cheaply and efficiently construct an instruction fine-tuning database with 4.4 million molecules related to chemical reaction and retrosynthesis for learning general synthesis-related knowledge, and introduce a dual-task learning strategy to simultaneously capture the forward and backward relationships when learning synthesis. 
}

\section{Method}

\textcolor{black}{Here, we propose ChemDual, a large language model with multi-scale tokenizer and dual-task learning. The overview of ChemDual is shown in Figure \ref{fig:ChemDual}, which is divided into three main modules. In the dataset construction module, we construct a 4.4M Molecule-Fragments database from 20M SMILES sequences with breaking of retrosynthetically interesting chemical substructures (BRICS) \cite{BRICS} (Section \ref{sec:dataset_construction}). Subsequently, in the module of multi-scale tokenizer, we extend the tokenizer the existing tokenizer to capture molecular information at different scales such as reaction and fragment (Section \ref{sec:multi-scale-tokenizer}). Finally, in the module of dual-task learning, we use dual-task learning on molecules and fragment as well as reaction and retrosynthesis to help LLM learn informative representation (Section \ref{sec:dual-task-learning}).}

% In this section, we present ChemDual, \textcolor{red}{which is implemented through a three-step strategy (Figure \ref{fig:ChemDual}). First, we construct a comprehensive molecular database by collecting and processing 2.2M SMILES sequences, which are then transformed into 4.4M molecular fine-tuning instructions. Second, we introduce a novel dual-task learning paradigm that leverages dual tasks in chemistry, including molecular fragmentation-recombination and reaction prediction-retrosynthesis. Finally, we detail the fine-tuning process of our LLM using both training optimization and generation strategies.}

\subsection{Dataset Construction}
\label{sec:dataset_construction}

% \subsubsection{Fragment \& Recombination}

\textcolor{black}{As shown in Figure \ref{fig:ChemDual}(a), we construct a chemical synthesis-related instruction dataset through data pre-processing, fragment-recombination operation, and template filling.}

\noindent \textbf{Data pre-processing}. \textcolor{black}{We collected 20M molecular SMILES sequences from the ChEMBL-34 database \cite{zdrazil2024chembl} and preprocessed these molecules according to three criteria: (i) removing duplicates; (ii) filtering out invalid molecules using RDKit \cite{bento2020open}; (iii) excluding molecules with weight greater than 1000. Next, we tokenized the molecular SMILES sequences and removed sequences with a token length of more than 512 to prevent exceeding the maximum token limit of LLM. Finally, we obtained a molecule library containing 2.2M SMILES sequences.}

% As shown in Figure \ref{fig:ChemDual}(a), we utilized the ChEMBL-34 \cite{zdrazil2024chembl} dataset, which contains 20M molecule SMILES sequences, and refined it by removing duplicates, filtering out invalid molecules using RDKit \cite{rdkit}, and excluding those with a molecular weight greater than 1000.
% Next, we tokenized the molecular SMILES sequences and removed sequences with token lengths exceeding 512 to prevent surpassing the maximum token limit of the LLM. 
% This process resulted in a final library of 2.2 million SMILES sequences.

\noindent \textbf{Fragment-recombination operation}. \textcolor{black}{Based on the molecular library, we adopt the BRICS algorithm to generate multiple fragments for the SMILES of the molecule and use these fragments to recombine back to the original SMILES. We regard the process of molecular fragmentation and recombination as a dual task. Meanwhile, we implement an adaptive fragmentation approach to alleviate memory overflow caused by excessive fragmentation when long SMILES sequences is processed by BRICS. Specifically, we adjust the number of fragments based on the length of molecule SMILES, and the maximum number of fragments is determined as follows:} %Specifically, to address memory overflow caused by excessive fragmentation when using BRICS on long SMILES sequences, we implemented an adaptive fragmentation approach that adjusts the number of fragmentations based on the molecule's length. The following formula determines the maximum number of fragmentations a SMILES sequence should be divided into:
\textcolor{black}{
\begin{equation}
n = 
\begin{cases} 
L, & \text{if } L < k; \\
\min\left(L, \left\lceil \frac{L}{k} \right\rceil^\alpha \right), & \text{otherwise.}
\end{cases}
\end{equation}
}
\textcolor{black}{where $L$ represents the length of the SMILES sequence, $\alpha$ denotes the elasticity factor to accommodate the capacities of different machine memory, and $k$ is the average length of SMILES sequences in molecule library. Interestingly, we observe that these fragments can be reassembled into a molecule, we term it as \textit{recombination}. In BRICS algorithm, dummy atoms are introduced at cleavage sites during the fragment labeling. We do not apply any additional processing to these dummy atoms during fragmentation, while these dummy atoms are removed and replaced with carbon (C) atoms to prevent the generation of invalid SMILES strings during recombination}.
%, which could potentially contaminate the model. 

\begin{figure}
    \centering
    \includegraphics[width=0.8\linewidth]{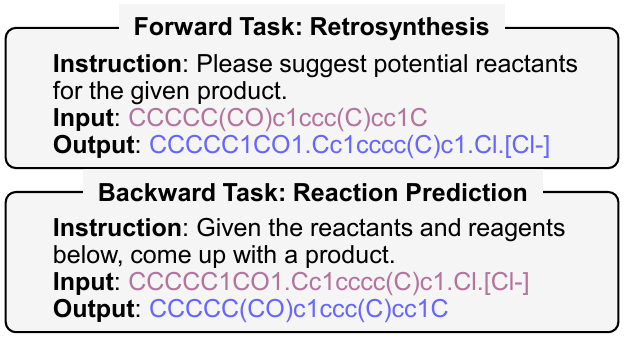}
    \caption{Examples of instruction of fine-tuning dataset.}
    \label{fig:instruction}
\end{figure}

\noindent \textbf{Template filling}. \textcolor{black}{To obtain more task-specific data and to adapt to the strong instruction-following abilities of LLMs, we design templates for chemical reaction and retrosysthesis prediction tasks. As shown in Figure \ref{fig:instruction}, in the forward task (i.e., retrosynthesis), we fix a general question format and then extract the molecular SMILES from molecule library to fill the description part of a predefined template, resulting in a natural question as instructions. The backward task (i.e., reaction prediction) is to take multiple products and corresponding reaction types (marked in green) as the question template. Through the entire pipeline, we finally construct a dataset containing 4.4M fine-tuning instructions for downstream prediction tasks.}

\subsection{\textcolor{black}{Multi-scale Tokenizer}}
\label{sec:multi-scale-tokenizer}

\textcolor{black}{BioT5 \cite{pei2023biot5} has demonstrated the advantages of using specialized tokenizer for LLMs to improve task-specific performance. As shown in Figure \ref{fig:ChemDual}(b), our proposed ChemDual further extends the tokenizer of LLaMA 3.1 by incorporating three additional types of tokenization strategies to handle the complexities of chemical data. 
The first category includes dummy atoms ({[1*], [2*], [3*], ..., [16*]}), which represent molecular fragments generated through the BRICS algorithm and encompass a total of 16 distinct dummy atom types. 
The second category consists of 180 functional groups commonly found in chemical structures, such as benzene rings, halogen atoms, and other commonly encountered functional groups in chemical structures. 
The third category introduces 4 special tokens, $<$BOF$>$ and $<$EOF$>$ for molecule and $<$BOM$>$ and $<$EOM$>$ for fragment, allowing the model to distinguish whether a sequence is a complete molecule or the corresponding fragment.
For a set of molecules or fragments, the input and output sequences are typically separated by a dot ("."). However, during tokenization, these sequences are further split into individual components using the special tokens discussed earlier. 
}

\subsection{Dual-task Learning}
\label{sec:dual-task-learning}

As illustrated in Figure \ref{fig:ChemDual}(c), we introduce a dual-task learning in both pre-training and instruction fine-tuning, to address the insufficient understanding of LLMs to the inherent structural information of SMILES.
By incorporating dual information into model training process, we aim to enable LLMs to capture more implicit chemical knowledge, thereby enhancing their understanding of chemical structure.
Specifically, our dual tasks include forward tasks and backward tasks, which use (fragmentation, recombination) and (reaction, retrosynthesis) to form the forward task and the backward task pair.
\textcolor{black}{This design is analogous to English–French translation and back-translation in NLP, where bidirectional learning reinforces semantic understanding.} The dual tasks can be represented mathematically as follows:
\begin{align}
    f_{\text{forward}} &: X \to Y, \quad (X, Y) \in \{(M, \mathcal{F}), (P, \mathcal{R})\}, \\
    f_{\text{backward}} &: Y \to X, \quad (Y, X) \in \{(\mathcal{F}, M), (\mathcal{R}, P)\},
\end{align}
\textcolor{black}{where $M$, $P$, $\mathcal{F}$ and $\mathcal{R}$ represent a molecule, the reaction product, the molecular fragments and reactants, respectively.}
% \begin{align}
%     f_{\text{forward}} &: X \to Y, \quad X \in \{M, P \}, \quad Y \in \{\mathcal{F}, \mathcal{R}\} \\
%     f_{\text{backward}} &: Y \to X, \quad Y \in \{\mathcal{F}, \mathcal{R}\}, \quad X \in \{M, P \}
% \end{align}

\textcolor{black}{The dual scenario can be represented by a more generalized function, we consider the joint probability of a data pair $(x, y)$, where $x \in X$ and $y \in Y$. Let $P(x)$ and $P(y)$ indicate the marginal distribution of $x$, $y$ respectively, $P(y|x; LLM)$ indicate the conditional probability of generating $y$ from $x$ using the LLM, and $P(x|y; LLM)$ indicate the conditional probability of generating $x$ from $y$ via LLM. Intuitively, we have:}
\begin{equation}
    P(x, y) = P(x)P(y|x; LLM) = P(y)P(x|y; LLM).
\end{equation}

% By employing this dual-task learning paradigm, the model is trained simultaneously on both tasks of molecular fragment and recombination, as well as reaction prediction and retrosynthesis. This duality not only helps LLM to understand the nature of chemical reactions more comprehensively, but also makes the model more flexible in reasoning when facing new chemical molecules or reactions.

% % \noindent\textbf{Pre-training.}
% % The pre-training phase aims to equip the model with foundational knowledge, enabling it to understand basic chemical structures and reactions. This phase involves training the LLM on a massive corpus of data consisting of molecular fragments, recombination pairs, and chemical reaction data. The goal is to enable the model to learn general representations of molecular components, structures, and reactions that can be fine-tuned later.

% % \noindent\textbf{Instruction Fine-Tuning.} The fine-tuning phase in the training of LLM involves adjusting the pre-trained model's parameters on a downstream task. During this process, the model is exposed to task-specific data, and the objective is to minimize the difference between the predictions from the model and the ground truth labels. 
\textcolor{black}{With the proposed dual-task learning framework, we can define the training mechanism, loss function, and optimization objective for LLM. Let $\mathcal{D} = {(x_i, y_i)}_{i=1}^{N}$ represent the dataset used for training, where $x_i$ represents the input sequence, and $y_i$ denotes the target sequence. Given an input sequence, the model is used to predict the probability distribution of possible output tokens, and its optimization goal is to maximize the likelihood of the correct sequence $y_i$. The probability distribution of the output vocabulary $\mathcal{V}$ of the model at each time step is formalized as follows:}
\begin{equation}
    P(y_i \mid x_i; \theta) = \text{softmax}(z_i),
\end{equation}
\textcolor{black}{where $z_i$ denotes the logits predicted by the model for the input sequence $x_i$, and $\theta$ represents the model parameters. 
We adopt cross-entropy loss as the objective function to measure the divergence between the predicted probability distribution and the true distribution (one-hot encoded label). The cross-entropy loss for a single training example $(x_i, y_i)$ is defined as follows:}
\begin{equation}
    \mathcal{L}(x_i, y_i; \theta) = -\sum_{j=1}^{|\mathcal{V}|} \mathbb{I}(y_i = j) \log P(y_i = j \mid x_i; \theta),
\end{equation}
\textcolor{black}{where $\mathbb{I}(y_i = j)$ is the indicator function that equals 1 when $y_i$ is the true label, and $0$ otherwise. 
% \textcolor{red}{The goal $\mathcal{L}_{a}$} is to minimize the average cross-entropy loss over the entire dataset:}
% \textcolor{red}{
% \begin{equation}
%     \mathcal{L}_{a}(x, y;\theta) = \frac{1}{N} \sum_{i=1}^{N} \mathcal{L}(x_i, y_i; \theta).
% \end{equation}
% }
}

\noindent \textbf{Pre-training.} 
\textcolor{black}{We perform pre-training on the fragment-recombination tasks. Specifically, we define fragment as the forward task and recombination as the backward task. 
The pre-training process of ChemDual is optimized as follows:}
\begin{equation}
\begin{aligned}
    \mathcal{L}_{pt}(x, y;\theta) &= \frac{1}{N} \sum_{i=1}^{N} 
    \mathcal{L}(x_i, y_i; \theta),
\end{aligned}
\end{equation}
\textcolor{black}{where $(x_i, y_i) \in \{(M, \mathcal{F}), (\mathcal{F}, M)\}$ evaluates the performance of fragment and recombination prediction on pre-training dataset. Dual-task learning enables the model to capture intrinsic relationships between molecular components, crucial for reaction and retrosynthesis prediction.}

\noindent \textbf{Instruction fine-tuning.} 
\textcolor{black}{We adapt the dual-task learning to fine-tuning more specific chemical synthesis tasks. Specifically, we regard retrosynthesis as the forward task, and reaction prediction is serves as the backward task. 
The fine-tuning process of ChemDual is optimized as follows:} 
\begin{equation}
\begin{aligned}
    \mathcal{L}_{ft}(x, y;\theta) &= \frac{1}{N} \sum_{i=1}^{N} 
    \mathcal{L}(x_i, y_i; \theta),
\end{aligned}
\end{equation}
\textcolor{black}{where $(x_i, y_i) \in \{(M, \mathcal{R}), (\mathcal{R}, M)\}$ accesses the performance of downstream tasks on the training dataset.}

% By employing these dual-task strategies in both pre-training and fine-tuning, the model learns to efficiently handle complex chemical data while capturing underlying relationships in molecular transformations and reactions. This process not only enhances the model's understanding of chemical structures and processes but also improves its performance on downstream tasks that involve molecular synthesis and reaction analysis.

% Given the large number of parameters in LLMs, fine-tuning all model parameters can be computationally expensive and inefficient. To address this, we employ Low-Rank Adaptation (LoRA) \cite{hu2021lora}.

% In LoRA, instead of updating the full weight matrix $W \in \mathbb{R}^{d \times k}$ in the model, we decompose the weight update into two smaller matrices $A \in \mathbb{R}^{d \times r}$ and $B \in \mathbb{R}^{r \times k}$, where $r \ll \min(d, k)$, making the parameter update more efficient. The modified weight matrix during training is then given by:
% \[
% W' = W + \Delta W = W + A B,
% \]
% where $A B$ represents the low-rank adaptation to the original weight matrix $W$. The matrices $A$ and $B$ are the only parameters trained during fine-tuning, drastically reducing the number of parameters and computational cost, while still allowing the model to adapt to the specific task. The fine-tuning process with LoRA optimizes these smaller matrices while keeping the original pre-trained weights $W$ frozen, which results in a more memory-efficient training.

\section{Experiments and Results}
\textcolor{black}{In this section, we first introduce compare our model ChemDual with baselines on reaction and retrosynthesis prediction. 
After that, we investigate the effectiveness of instruction dataset and fine-tuning strategy in our model. Finally, we conduct the visualization analysis and case study via molecular docking.}
\textcolor{black}{Details of the experimental settings are provided in Appendix \ref{app:D}.
}
%In this section, we subject our proposed methodologies to a rigorous evaluation, focusing on the evaluation metrics of chemical structures and NLP, to address two fundamental research questions. 

\begin{table*}[htbp]
% \caption{Performance comparison for reaction and retrosynthesis prediction on Mol-Instruction dataset. [*] means the results are taken from \cite{mol-instruction}. The best and \underline{suboptimal} results are shown in bold and underline.}
\caption[Performance comparison for reaction and retrosynthesis prediction]{
Performance comparison for reaction and retrosynthesis prediction on the Mol-Instruction dataset. [*] indicates the results are taken from Mol-Instruction~\cite{mol-instruction}. The \textbf{best} and \underline{suboptimal} results are shown in bold and underline.
}
\label{tab:dual2}
\begin{center}
\resizebox{\textwidth}{!}{%
\begin{tabular}{lccccccc}

\toprule
\multicolumn{1}{c}{\bf Model} & \multicolumn{1}{c}{\bf EXACT↑} & \multicolumn{1}{c}{\bf BLEU↑} & \multicolumn{1}{c}{\bf LEVENSHTEIN↓} & \multicolumn{1}{c}{\bf RDK FTS↑} & \multicolumn{1}{c}{\bf MACCS FTS↑} & \multicolumn{1}{c}{\bf MORGAN FTS↑} & \multicolumn{1}{c}{\bf VALIDITY↑} \\ \midrule

\rowcolor{gray!15} \multicolumn{8}{l}{Reaction Prediction} \\
Alpaca* & 0.000 & 0.065 & 41.989 & 0.004 & 0.024 & 0.008 & 0.138 \\
Baize* & 0.000 & 0.044 & 41.500 & 0.004 & 0.025 & 0.009 & 0.097 \\
ChatGLM* & 0.000 & 0.183 & 40.008 & 0.050 & 0.100 & 0.044 & 0.108 \\
LLaMA* & 0.000 & 0.020 & 42.002 & 0.001 & 0.002 & 0.001 & 0.039 \\
Vicuna* & 0.000 & 0.057 & 41.690 & 0.007 & 0.016 & 0.006 & 0.059 \\
Galactica* & 0.000 & 0.468 & 35.021 & 0.156 & 0.257 & 0.097 & 0.946 \\
Text+Chem T5* & 0.239 & 0.782 & 20.413 & 0.705 & 0.789 & 0.652 & 0.762 \\
Mol-Instruction* & 0.503 & 0.883 & 13.41 & 0.756 & 0.863 & 0.708 & \textbf{1.000} \\
BioT5+ & \underline{0.864} & \textbf{0.993} & \underline{3.403} & \underline{0.949} & \underline{0.975} & \underline{0.935} & \textbf{1.000} \\
% ChemDual$_{\textit{w/o\_pre}}$   & {0.821} & {0.968} & \underline{2.532} & \underline{0.954} & \underline{0.976} & \underline{0.947} & 0.988 \\
ChemDual & \textbf{0.869} & \underline{0.991} & \textbf{2.099} & \textbf{0.964} & \textbf{0.980} & \textbf{0.956} & \underline{0.996} \\ \toprule

\rowcolor{gray!15} \multicolumn{8}{l}{Retrosynthesis} \\
Alpaca* & 0.000 & 0.063 & 46.915 & 0.005 & 0.023 & 0.007 & 0.160 \\
Baize* & 0.000 & 0.095 & 44.714 & 0.025 & 0.050 & 0.023 & 0.112 \\
ChatGLM* & 0.000 & 0.117 & 48.365 & 0.056 & 0.075 & 0.043 & 0.046 \\
LLaMA* & 0.000 & 0.036 & 46.844 & 0.018 & 0.029 & 0.017 & 0.010 \\
Vicuna* & 0.000 & 0.057 & 46.877 & 0.025 & 0.030 & 0.021 & 0.017 \\
Galactica* & 0.000 & 0.452 & 34.940 & 0.167 & 0.274 & 0.134 & 0.986 \\
Text+Chem T5* & 0.141 & 0.765 & 24.043 & 0.685 & 0.765 & 0.585 & 0.698 \\
Mol-Instruction* & 0.333 & 0.842 & 17.642 & 0.704 & 0.815 & 0.646 & \textbf{1.000} \\
BioT5+ & \underline{0.642} & \underline{0.969} & \underline{6.710} & \underline{0.897} & \underline{0.930} & \underline{0.866} & \textbf{1.000} \\
% ChemDual$_{\textit{w/o\_pre}}$            & {0.587} & {0.868} & {11.026} & {0.848} & {0.897} & {0.812} & 0.985 \\
ChemDual & \textbf{0.670} & \textbf{0.976} & \textbf{6.516} & \textbf{0.901} & \textbf{0.933} & \textbf{0.893} & \underline{0.995} \\
\bottomrule

\end{tabular}%
}
\end{center}
\end{table*}

\begin{table}[h!]
\centering
\caption{\textcolor{black}{Performance comparison on USPTO-50K dataset, [*] means ChemDual adopts the Retroformer’s transformer module.}}
\resizebox{0.45\textwidth}{!}{%
\begin{tabular}{lcccc}
\toprule
\textbf{Model} & \textbf{Top-1 (\%)} & \textbf{Top-3 (\%)} & \textbf{Top-5 (\%)} & \textbf{Top-10 (\%)} \\
\midrule
Transformer   & 42.40 & 58.60 & \underline{63.80} & 67.70 \\
BioT5+        & \underline{44.40} & \underline{59.56} & 61.32 & \underline{73.43} \\
InstructMol   & 30.15 & 51.72 & 57.12 & 64.91 \\
\textbf{ChemDual} & \textbf{46.25} & \textbf{60.14} & \textbf{66.95} & \textbf{77.42} \\
\midrule
Retroformer   & \underline{47.89} & \underline{62.87} & \underline{66.59} & \underline{70.68} \\
\textbf{ChemDual*} & \textbf{49.95} & \textbf{67.67} & \textbf{70.52} & \textbf{78.31} \\
\midrule
Improvement(\%) & 2.06 & 4.79 & 3.93 & 7.63 \\
\bottomrule
\end{tabular}}
\label{tab:topk_accuracy}
\end{table}
%\vspace{-2em}

% In this section, we subject our proposed methodologies to a rigorous evaluation, focusing on the evaluation metrics of chemical structures and NLP, in order to address two fundamental research questions that guide our investigation. For the Issue \ref{issue:smiles}, we examine whether the constructed 4.4M molecule splitting and recombination dataset can enhance the LLM’s understanding of chemical structural information. As for the Issue \ref{issue:dual}, we explore whether the dual-form fine-tuning paradigm proves effective for LLM in terms of both chemical structure understanding and NLP metrics. Our approach includes integrating the dataset and applying multi-step fine-tuning to assess any improvements in the model's performance.

% First, we examine whether our constructed 4.4M molecule splitting and recombination dataset can enhance the LLM's understanding of chemical structural information, given that existing biochemical LLMs often struggle with understanding SMILES notation due to their reliance on simple sequence pre-training. Second, we investigate the effectiveness of our dual-form fine-tuning paradigm in improving both chemical structure understanding and NLP metrics, considering that existing models lack a comprehensive dual-task learning strategy to capture the inter-dependencies between forward and backward processes. Our approach includes integrating the dataset and applying multi-step fine-tuning to assess any improvements in the model's performance.

\subsection{Experiment Setup}

%We provide detailed information on the hyperparameter settings used for pre-training and fine-tuning LLaMA 3.1.

\noindent \textbf{Dataset.}
\textcolor{black}{We use Mol-Instruction and ChemLLMBench dataset to evaluate ChemDual on reaction and retrosynthesis prediction. In addition, we evaluate the retrosynthesis task on the USPTO-50 dataset.
The Mol-Instruction consists of 282,304 molecules paired instructions describing their reactions, and each sample is represented by a tuple (input, output), where the input could be either a set of reactants or a molecule, and the output corresponds to the predicted products or reactants. %We follow the original dataset split. %$, we train and test on their reaction prediction and retrosynthesis tasks.
ChemLLMBench consists of a series of chemistry-related tasks. In this work, we evaluate ChemDual on the reaction and retrosynthesis prediction tasks within ChemLLMBench. We follow the same test set as used in ChemDFM \cite{chemdfm}.
The USPTO-50k dataset contains 50,016 reaction examples that are grouped into 10 reaction classes, and each sample represents a synthetic reaction step with corresponding reactants and products. We keep the same data split as Graph2SMILES \cite{tu2022permutation}.
}

%\textcolor{blue}{We use two benchmarking datasets to evaluate the performance of our proposed method. Specifically, the USPTO50k dataset contains 50,016 reaction examples, each representing a synthetic reaction step with corresponding reactants and products. It covers a wide variety of chemical reactions, including those found in organic synthesis, and is widely used for training and testing models on retrosynthesis task. On the other hand, the Mol-Instruction dataset consists of 282,304 molecules paired with instructions describing their reactions. This dataset is designed for instruction-based fine-tuning and each sample in these datasets is represented as a tuple (input,output), where input could be either a set of reactants or a molecule, and output corresponds to the predicted products or reactants.}

\noindent \textcolor{black}{\textbf{Baseline Models.} We compare ChemDual against a variety of baselines which can be categorized as follows:}
\begin{itemize}
    \item \textbf{General LLMs:} Alpaca \cite{alpaca}, Baize \cite{baize}, ChatGLM \cite{glm2024chatglm}, LLaMA \cite{touvron2023llamaopenefficientfoundation} and Vicuna \cite{vicuna2023}, 
    %they are all trained LLMs that demonstrate strong performance in general-purpose tasks.
    \item \textbf{Specialized models:} Galactica \cite{galactica2022}, Text+Chem T5 \cite{text+chemt5}, Mol-Instruction, BioT5+ \cite{pei2024biot5+}, InstructMol \cite{cao2023instructmol}, Retroformer \cite{wan2022retroformer}. 
    %all of which are tailored models that demonstrate strong performance in specific domains.
\end{itemize}

\noindent \textbf{Evaluation metrics.} \textcolor{black}{For reaction and retrosynthesis tasks, we employ various metrics to evaluate the effectiveness of ChemDual and all baselines, including EXACT, BLEU \cite{papineni2002bleu}, LEVENSHTEIN \cite{levenshtein1966binary}, three molecular fingerprints (FTS) similarity scores including RDK \cite{schneider2015get}, MACCS \cite{durant2002reoptimization}, MORGAN \cite{rogers2010extended} and VALIDITY score refers to whether the SMILES can be successfully processed by RDKit. Additionaly, we further adopt Top-k accuracy as evaluation metric on retrosynthesis prediction.}

\begin{figure*}[tbp]
\begin{center}
    \includegraphics[width=1\textwidth]{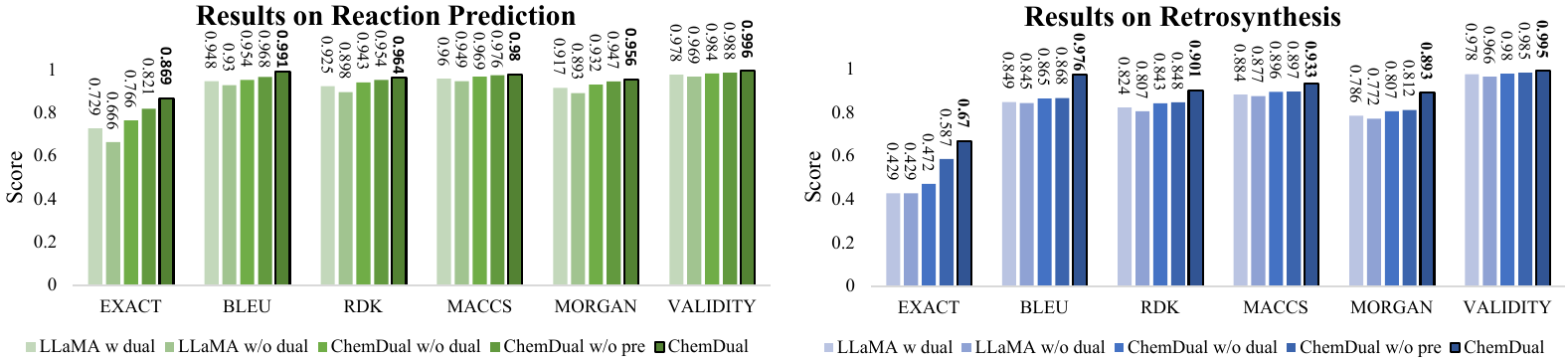}
    \caption{\textcolor{black}{Comparison of ablation experiments for reaction prediction and retrosynthesis prediction on Mol-Instruction.}}
    \label{fig:ablation-comparison}
\end{center}
\end{figure*}

\subsection{Comparison Results}
Table \ref{tab:dual2} shows the comparison results for reaction and retrosynthesis prediction on Mol-Instruction.
\textcolor{black}{
Detailed results on ChemLLMBench can be found in Appendix \ref{app:A}.
}
% \subsubsection{Forward Reaction Prediction \& Retrosynthesis}
%In this section, we compare the reaction prediction performance of ChemDual with the baselines on the Mol-Instruction dataset, as well as its retrosynthesis performance on both the Mol-Instruction and USPTO-50k datasets.

\noindent \textbf{Results on reaction prediction}.
\textcolor{black}{As shown in the top of Table \ref{tab:dual2}, we observe that ChemDual achieves a substantial improvement with the highest EXACT score of 0.869 and the lowest LEVENSHTEIN distance of 2.099.
%outperforming the second-best model, BioT5+, by 0.005 in EXACT, and 1.304 in LEVENSHTEIN, respectively. 
Moreover, ChemDual consistently outperformed all baselines across multiple structure similarity metrics such as RDK, MACCS, and MORGAN FTS, with a promising scores of 0.964, 0.980, and 0.956, respectively. This indicates that our proposed ChemDual can perform highly chemical relevant predictions compared to other methods. However, ChemDual achieves competitive VALIDITY score with Mol-Instruction and BioT5+, the slight gap may be attributed to the reason that ChemDual adopt SMILES representations as input while Mol-Instruction and BioT5+ employ the unique SELFIES format.}

%As indicated by the performance in the top of Table \ref{tab:dual2}, we found that ChemDual achieved a substantial improvement, obtaining the highest EXACT score of 0.869 and the lowest LEVENSHTEIN distance of 2.099, outperforming the second-best model, BioT5+, by 0.005, and 1.304, respectively. Moreover, ChemDual consistently outperformed baselines in molecular similarity metrics such as RDK, MACCS, and MORGAN FTS, achieving scores of 0.964, 0.980, and 0.956, respectively. This indicates that ChemDual can generate highly accurate and chemically relevant predictions compared to other methods. However ChemDual achieved a VALIDITY score of 0.996, very close to the perfect VALIDITY of 1.000. The slight gap in VALIDITY can be attributed to the use of SMILES representations in ChemDual, whereas MOL-Instruction and BioT5+ employed SELFIES.

\noindent \textbf{\textcolor{black}{Results on retrosynthesis prediction}.} 
\textcolor{black}{The bottom of Table \ref{tab:dual2} shows that ChemDual achieves the best performance of 0.670 in EXACT, 0.976 in BLEU, and 6.516 in LEVENSHTEIN distance, respectively, with comparison to other models, including the strong baseline BioT5+. Furthermore, ChemDual also surpasses BioT5+ by 0.4\%, 0.3\%, and 2.7\% in terms of RDK, MACCS, and MORGAN FTS, respectively, which implies that ChemDual performs the ability to generate accurate and chemical retrosynthesis predictions.} 

\textcolor{black}{We further evaluate the retrosynthesis performance of ChemDual in USPTO-50K dataset. As shown in Table \ref{tab:topk_accuracy}, ChemDual achieves the best performance in all top-k accuracy, which demonstrates its robustness and scalability across datasets. Specifically, ChemDual outperforms BioT5+ and other baseline models at least by 1.85\%, 0.58\%, 3.15\%, and 3.99\% in terms of Top-k (k=1, 3, 5, and 10), respectively, Compared with Retroformer, the fine-tuned ChemDual* consistently makes improvements of 2.06\% in Top-1, 4.79\% in Top-3, 3.93\% in Top-5, and 7.63\% in Top-10, respectively.}

%Compared with fine-tuned ChemDual*, it further improves to 49.95\%, surpassing Retroformer (47.89\%) by 2.06\% in top-1 accuracy. This trend continues across top-3, top-5, and top-10 accuracies, where ChemDual consistently achieves superior results, with improvements of 4.79\%, 3.93\%, and 7.63\%, respectively, over the best baselines.
% General-purpose models like Alpaca, Baize, and LLaMA demonstrate significantly lower performance across all metrics. These models are not specialized for chemical tasks, which likely explains their inability to generate precise or chemically valid results.

\noindent\textbf{Results on fragment and recombination.} \textcolor{black}{To investigate whether ChemDual accomplish the molecular fragment and recombination tasks, we evaluate our proposed model using 1,000 samples from our constructed instruction test set. The details of results refer to Appendix \ref{app:B}.}

\subsection{Ablation study}

\textcolor{black}{To investigate the effectiveness of each component to our model, we consider the following variants of ChemDual. Figure \ref{fig:ablation-comparison} shows the results of ChemDual and its variants on reaction and retrosysthesis prediction.} 

\begin{itemize}
    \item \textcolor{black}{\textit{LLaMA with dual-task learning} (w dual) removes the datasets related to molecule fragments and recombination, and we trained the LLaMA 3.1 only using Mol-Instruction dataset.}
    \item \textcolor{black}{\textit{ChemDual/LLaMA without dual-task learning} (w/o dual) deletes the datasets related to reaction prediction in retrosynthes prediction task, and the datasets related to retrosynthes is prediction are deleted reaction prediction task, respectively.}
    \item \textcolor{black}{\textit{ChemDual without pre-training} (w/o pre) directly fine-tunes the LLaMA 3.1 on Mol-Instruction and our constructed instruction datasets with a ratio of 1:1.}
    % \item \textit{ChemDual } first pre-trained the model on the 4.4M ChemDual-Instruction dataset in the initial phase, followed by instruction fine-tuning on Mol-Instruction. 
\end{itemize}

%We conducted the ablation study for reaction prediction and retrosynthesis on the LLaMA 3.1 from two perspectives: datasets ablation and fine-tuning strategy ablation to examine the impact on model performance of the large-scale dataset and the dual-task fine-tuning paradigm to help LLMs understand chemical structure information. The results, summarized in Figure \ref{fig:ablation-comparison}, underscore the effectiveness of our approach.
%disrupted the dual-task learning strategy by systematically removing key components. We use "w/o dual" to indicate the

\begin{figure}[h]
    \centering
    \includegraphics[width=1\linewidth]{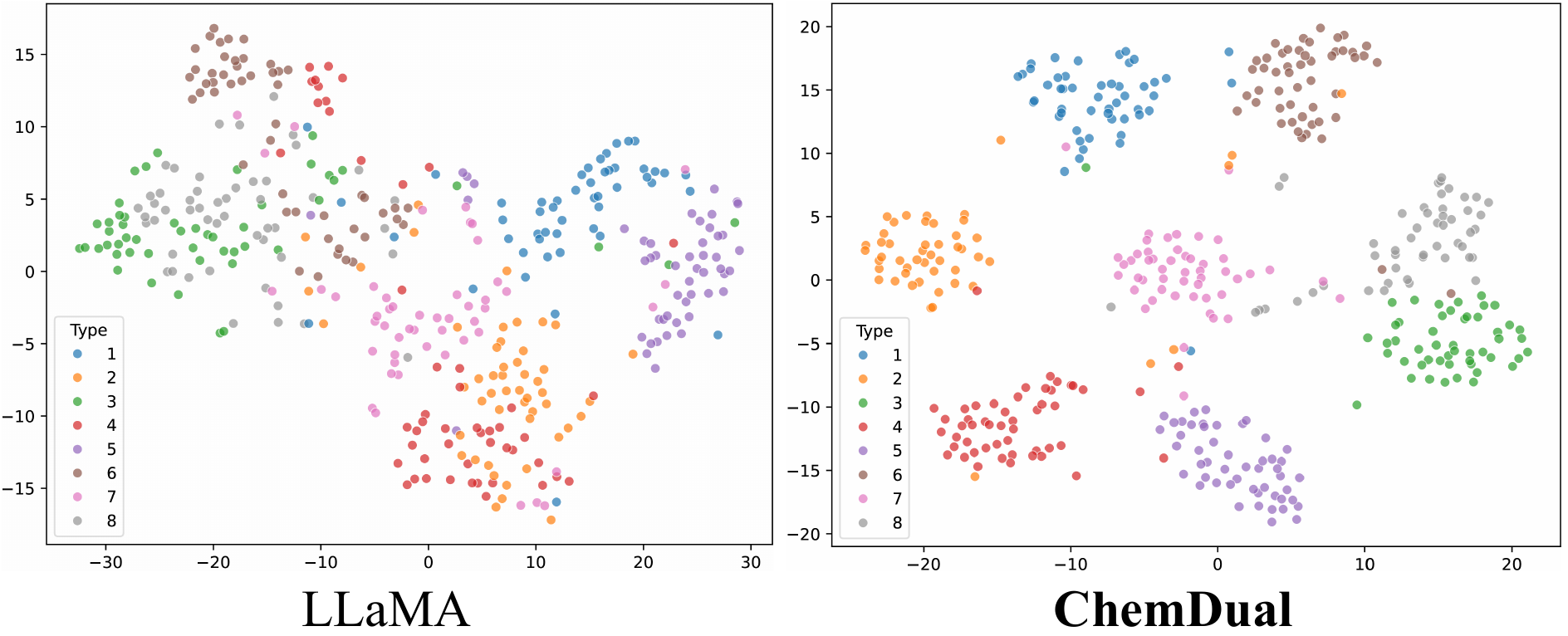}
    \caption{Visualizations of LLaMa and ChemDual using the t-SNE.}
    \label{fig:visualization}
\end{figure}

\begin{figure*}[t]
    \centering
    \includegraphics[width=0.88\linewidth]{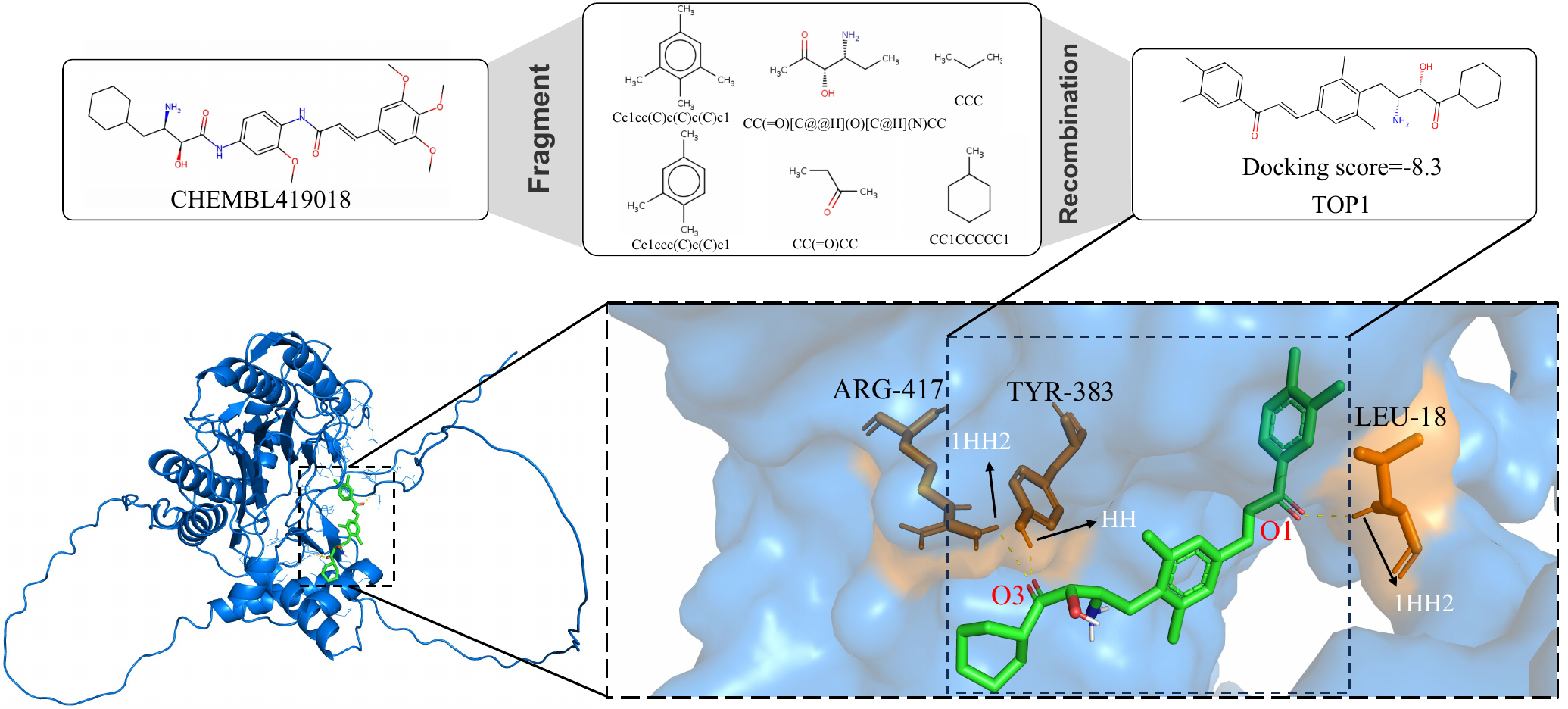}
    \caption{\textcolor{black}{Visualization of the molecular docking complex between the top1 molecule generated by ChemDual and the protein. The binding pose shows three significant hydrogen bond interactions: the O3 atom of the ligand forms hydrogen bonds with the 1HH2 atom of ARG417 and the HH atom of TYR383, while the O1 atom of the ligand establishes another hydrogen bond with the NH group of LEU18.}}
    \label{fig:docking}
\end{figure*}

\noindent\textcolor{black}{\textbf{Dataset ablation.}} 
% In the dataset ablation study, we removed the datasets related to molecule fragments and recombination, training the model on LLaMA 3.1 using only the paired datasets for reaction prediction and retrosynthesis, which we refer to as "LLaMA w dual."
% For the "ChemDual", we first pre-trained the model on the 4.4M ChemDual-Instruction dataset in the initial phase, followed by instruction fine-tuning on Mol-Instruction. 
\textcolor{black}{ChemDual outperforms the LLaMA (w dual) on all metrics, which demonstrates the importance of leveraging our constructed instruction dataset for enhancing predictive performance.
}

\noindent
\textbf{Pre-training ablation.} 
% In pre-training ablation, we design two sets of reference experiments. "ChemDual w/o pre" refers to instruction fine-tuning the model directly using on a 1:1 mixture of the Mol-Instruction and ChemDual-Instruction datasets, without any pre-training.
\textcolor{black}{The comparison of ChemDual and ChemDual (w/o pre) reveals a consistent improvement on model performance for the pre-trained model (i.e., ChemDual) across reaction and retrosynthesis prediction tasks. This highlights the critical role of pre-training in capturing the intrinsic relationships between molecular component.}

% Specifically, our 4.4M ChemDual-Instruction dataset provided substantial information about chemical bonds and structures, resulting in performance gains of 13\% in reaction prediction and 9\% in retrosynthesis tasks. These findings demonstrate that training on our comprehensive dataset significantly enhances the model's understanding of SMILES semantic information, enabling more accurate chemical predictions.

\noindent \textbf{Dual-task learning ablation.} 
% We disrupted the dual-task learning strategy by systematically removing key components. We use "w/o dual" to indicate the removal of the reaction prediction dataset in reaction prediction task and removal of the retrosynthesis dataset in retrosynthesis task. 
\textcolor{black}{Dual-task method (i.e., LLaMA (w dual)) performs better than single-task model (i.e., ChemDual/LLaMA (w/o dual)), which illustrates that the model is fine-tuned by the dual-task paradigm, either independently or with the inclusion of molecular fragment and recombination datasets, consistently outperformed than that is trained on a single dataset. These results underscore the effectiveness of the dual-task learning approach, which promotes the synergistic learning across tasks and improves the capabilities of model on chemical reasoning .}

\textcolor{black}{The ablation studies further validate the critical role of instruction dataset and dual-task learning paradigm in enhancing the performance of general LLM on reaction and retrosynthesis prediction tasks. We also highlight that a well-designed pre-training strategy allows LLM to better grasp the semantics of chemical structures, enabling them to make more precise predictions across multiple downstream tasks.
}

% From Table \ref{table3}, we can draw the following conclusions: (i) Both approaches, whether fine-tuning LLaMA 3.1 using the dual paradigm alone or incorporating the molecular fragment and recombination datasets into the dual paradigm fine-tuning, outperform the experiments that only use a single dataset. This result indicates that the dual training and fine-tuning paradigm is effective. (ii) When comparing the models "LLaMA-trained", "ChemDual$_{\textit{w/o\_pre}}$" and "ChemDual," we observe a trend of improving performance. This suggests that within our constructed 4.4M ChemDual-Instruction dataset, LLM is able to learn sufficient information about chemical bonds and structures, leading to respective exactitude improvements of 13\% in forward reaction prediction and 9\% in retrosynthesis tasks. This result demonstrates that training on our dataset indeed enhances the model’s understanding of SMILES semantic information.

\section{Analysis}
\noindent \textbf{Visualization.} 
\textcolor{black}{To better understand the superiority of ChemDual over the general LLM (i.e., LLaMa), we employ t-SNE \cite{t-sne} to visualize the learned representations of input prompt. Since there are dozens of reaction types, we select 8 reaction types and each with 50 sample data for visualization. Figure \ref{fig:visualization} shows the visualization on USPTO-50K dataset. We clearly observe that input prompts are more tightly clustered in ChemDual compared with LLaMa, which implies that ChemDual can learn more contextual information by effectively capturing the chemical synthesis-related knowledge from our newly constructed instruction dataset. It is worth noting that more compact clusters in ChemDual on 8 reaction types can well illustrate the excellent ability of ChemDual on retrosynthesis prediction.}

%For a more intuitive comparison, we employ t-SNE to visualize the type of reaction representations generated by the last layer of ChemDual. Specifically, we select 50 sample points for each of the 8 reaction types from the USPTO-50K test set. To extract these representations, we focus on the hidden states corresponding to the last token of each sequence, which is the \textbf{\texttt{<|end\_of\_text|>}} token output. This ensures that the representations capture the complete contextual information of the entire sequence. As shown in Figure \ref{fig:visualization}, the representations generated by ChemDual are distinctly separated according to the type of reaction. This distinction highlights ChemDual’s superior capability in capturing reaction-specific features.

\noindent\textbf{Molecule recombination and docking.} 
\textcolor{black}{We conduct a case study to investigate the ability of our proposed ChemDual on molecular recombination. Traditional methods struggle with balancing chemical validity and structural diversity. Here we further explore how our constructed instruction dataset addresses this challenge by generating molecules with high chemical similarity. 
\textcolor{black}{Specifically, we select the compound (CHEMBL419018) from the available CHEMBL database. By fragmenting the compound with BRICS, we obtain multiple fragments as shown in Figure \ref{fig:docking}. Based on these fragments, we are able to generate novel compounds (the details refers to Figure \ref{fig:case1} in Appendix \ref{app:C}) via ChemDual.} These generated compounds inherit characteristics from their original fragments and are highly similar to the original molecules in molecular fingerprints, which demonstrates the ability of ChemDual to maintain key chemical features while generating valid molecular structures, ensuring both chemical integrity and similarity during the recombination process.
}

\textcolor{black}{Methionine aminopeptidase 2 (MAP2) is an enzyme of important biological significance, as it plays a key role in angiogenesis and tumor progression. It has been recognized as a therapeutic target, with several inhibitors demonstrating promising anti-cancer and anti-obesity activities in both preclinical and clinical studies. Therefore, we use the MAP2 as target protein (CHEMBL419018 in the CHEMBL34 dataset) and conduct docking experiments between the protein and the generated compounds to evaluate binding affinities. As shown in Figure \ref{fig:case1}, the binding affinity values of the generated compounds range from -6.3 to -8.4 kcal/mol, which indicates that the molecules generated by ChemDual have good binding interactions with the target protein. } 
%which suggests that ChemDual-Instruction's generated molecules consistently exhibit favorable binding interactions with the target protein, supporting the model's ability to generate compounds with both chemical coherence and functional relevance.
To gain deeper insights, we use AutoDock \cite{Morris2009AutoDock} to further visualize the docking result and binding interactions of the first generated compound in Figure \ref{fig:docking}. The visualization highlights three key hydrogen bonds formed by the ligand: the first is between the O3 atom of the ligand and the 1HH2 atom of the ARG-417 residue; the second is between the O3 atom of the ligand and the HH atom of the TYR-383 residue; and the third is between the O1 atom of the ligand and the NH group of the LEU18 residue. These interactions underscore the stability of the ligand within the protein's active site, demonstrating that ChemDual can not only recombine structurally similar molecules but also yield compounds with stable, specific binding characteristics in target interactions.

% \subsection{Integrating retrosynthesis for Practical Application}

% Although ChemDual enables the generation of numerous combination-based compounds, the challenge of synthesizing these compounds in practice remains substantial. Practical synthesis requires not only theoretical design but also a feasible synthesis pathway to transform the generated molecular structures into real, usable compounds. To address this critical gap, we developed a downstream synthesis instruction module grounded in training with molecule fragment and recombination, specifically aimed at facilitating the synthesis of complex, recombined molecules. This module equips ChemDual to suggest step-by-step retrosynthetic pathways, tailored to guide the practical synthesis of combination compounds.

% As illustrated in Figure \ref{fig:case2}, we screened five common small molecules from the ZINC database for recombination. Subsequently, we proceeded to generate based on the given prompts for the retrosynthesis task. ChemDual is now capable not only of generating compounds that satisfy molecular efficacy requirements but also of supporting retrosynthesis to enable real-world production of these structures. By providing retrosynthetic instructions, ChemDual allows researchers to move from computational generation to actionable synthesis steps, thereby bridging the gap between theoretical recombination and experimental realization. This advancement enhances ChemDual's utility in drug discovery, offering comprehensive support in both innovative molecular design and practical chemical synthesis.

\section{Conclusion}
\textcolor{black}{
% In this study, we propose an enhanced large language model (LLM) framework based on LLaMA, called ChemDual, for prediction of chemical reaction and retrosynthesis. ChemDual first constructs a database with 4.4 million molecules for alleviating the lack of a large-scale chemical synthesis-related instruction dataset. Subsequently, ChemDual uses a multi-scale tokenizer to learn multiple scales of structural information, and introduce a dual-task learning strategy to jointly optimize the process of recombination and fragmentation as well as the tasks between reaction and retrosynthesis prediction, which solves the challenge of ignoring the correlation between them.
% Through extensive experiments, we demonstrate the superior performance of ChemDual on Mol-Instruction and USPTO-50K, which outperforms a variety of existing methods. In addition, experiments on molecular docking show that the proposed ChemDual can generate compounds with high affinity to the target protein, indicating the strong potential of ChemDual for molecular design.
In this study, we propose ChemDual, an enhanced LLaMA-based LLM for chemical reaction and retrosynthesis prediction. It constructs a 4.4-million-molecule database to alleviate the lack of a large-scale chemical synthesis-related instruction dataset, employs a multi-scale tokenizer to capture structural information, and introduces a dual-task learning strategy to jointly optimize the process of recombination and fragmentation as well as the tasks between reaction and retrosynthesis prediction.
Experiments on Mol-Instruction and USPTO-50K demonstrate that ChemDual outperforms existing methods, while molecular docking results show its ability to generate compounds with strong affinity for target proteins, highlighting its potential in molecular design.
}

\bibliographystyle{named}
\bibliography{references}

% \appendix
% \include{appendix}
% \clearpage

% %% The file named.bst is a bibliography style file for BibTeX 0.99c
% \bibliographystyle{named}
% %\bibliographystyle{plain}
% \bibliography{ijcai25}

\clearpage

\appendix   %仅一个附录时用appendix，否则\appendices
\onecolumn
% \setcounter{table}{0}   %从0开始编号，显示出来表会A1开始编号
% \setcounter{figure}{0}
% \setcounter{section}{0}
% \setcounter{equation}{0}
%定义编号格式，在数字序号前加字符“A"
% \renewcommand{\thetable}{A\arabic{table}}
% \renewcommand{\thefigure}{A\arabic{figure}}
% \renewcommand{\thesection}{A\arabic{section}}
% \renewcommand{\theequation}{A\arabic{equation}}

\begin{center}
    \Huge Appendix
\end{center}

\section{Experiment setting}
\label{app:D}
% \noindent \textbf{Experiment setting.}
\textcolor{black}{
ChemDual is implemented using Pytorch v2.4.1 and CUDA 12.1. In the training phase, all experiments are conducted on the same machine with Intel Xeon(R) Platinum 8352V CPU @ 2.10GHz and 4 GPUs (NVIDIA RTX 4090 GPUs 24G) for a total of 212 GPU hours and processed 852,518,047 tokens. 
We use the AdamW optimizer to facilitate the optimization process. Each GPU was assigned a batch size of 2, with gradient accumulation steps set to 8.}
The pre-training phase was conducted for 1 epoch with a learning rate of 5e-5, which enable stable and gradual learning during the initial stages. The fine-tuning phase was carried out over 2 epochs with a learning rate of 1e-4.
\textcolor{black}{
In the inference phase, the model was quantized to INT4 and ran on a single 4090 GPU at a speed of 98.16 tokens/s.
}

\section{Experiment on ChemLLMBench}
\label{app:A}
\textcolor{black}{
As shown in Table \ref{tab:comparison}, ChemDual demonstrates superior performance on both reaction prediction and retrosynthesis tasks within the ChemLLMBench. Specifically, ChemDual achieves the highest accuracy of 67.0 in reaction prediction, outperforming ChemDFM (49.0), BioT5+ (9.0), and Mol-Instruction (4.5) by a substantial margin. Similarly, in retrosynthesis prediction, ChemDual achieves an accuracy of 33.0, which also surpasses other baselines, indicating the model’s improved reasoning ability on backward chemical tasks. While maintaining competitive validity scores across both tasks, ChemDual exhibits a strong balance between correctness and chemical plausibility, highlighting the effectiveness of the dual-task learning framework in enhancing structural understanding and generalization in complex chemical scenarios.
}
\begin{table}[h]
    \caption{Performance comparison on ChemLLMBench.}
    \label{tab:comparison}
    \begin{center}
    \scriptsize
    
    \begin{tabular}{l l c c}
        \toprule
        \textbf{Task} & \textbf{Model} & \textbf{Accuracy} & \textbf{Validity} \\
        \midrule
        \textbf{Reaction Prediction} & Mol-Instruction \cite{mol-instruction} & 4.5 & \textbf{100.0} \\
        & BioT5+ \cite{pei2024biot5+} & \underline{9.0} & \textbf{100.0} \\
        & ChemDFM \cite{chemdfm} & \underline{49.0} & \textbf{98.0} \\
        & ChemDual & \textbf{67.0} & \underline{99.0} \\
        \midrule
        \textbf{Retrosynthesis} &  Mol-Instruction \cite{mol-instruction} & 9.0 & \textbf{100.0} \\
        & BioT5+ \cite{pei2024biot5+} & \underline{26.0} & \textbf{100.0} \\
        & ChemDFM \cite{chemdfm} & \underline{12.0} & \textbf{91.0} \\
        & ChemDual & \textbf{33.0} & \underline{97.7} \\
        \bottomrule
    \end{tabular}
    \end{center}
\end{table}

\section{Experiment on fragment and recombination}
\label{app:B}
\begin{figure*}[b]
\centering
\begin{subfigure}[t]{0.45\textwidth}
    \centering
    \includegraphics[width=\linewidth]{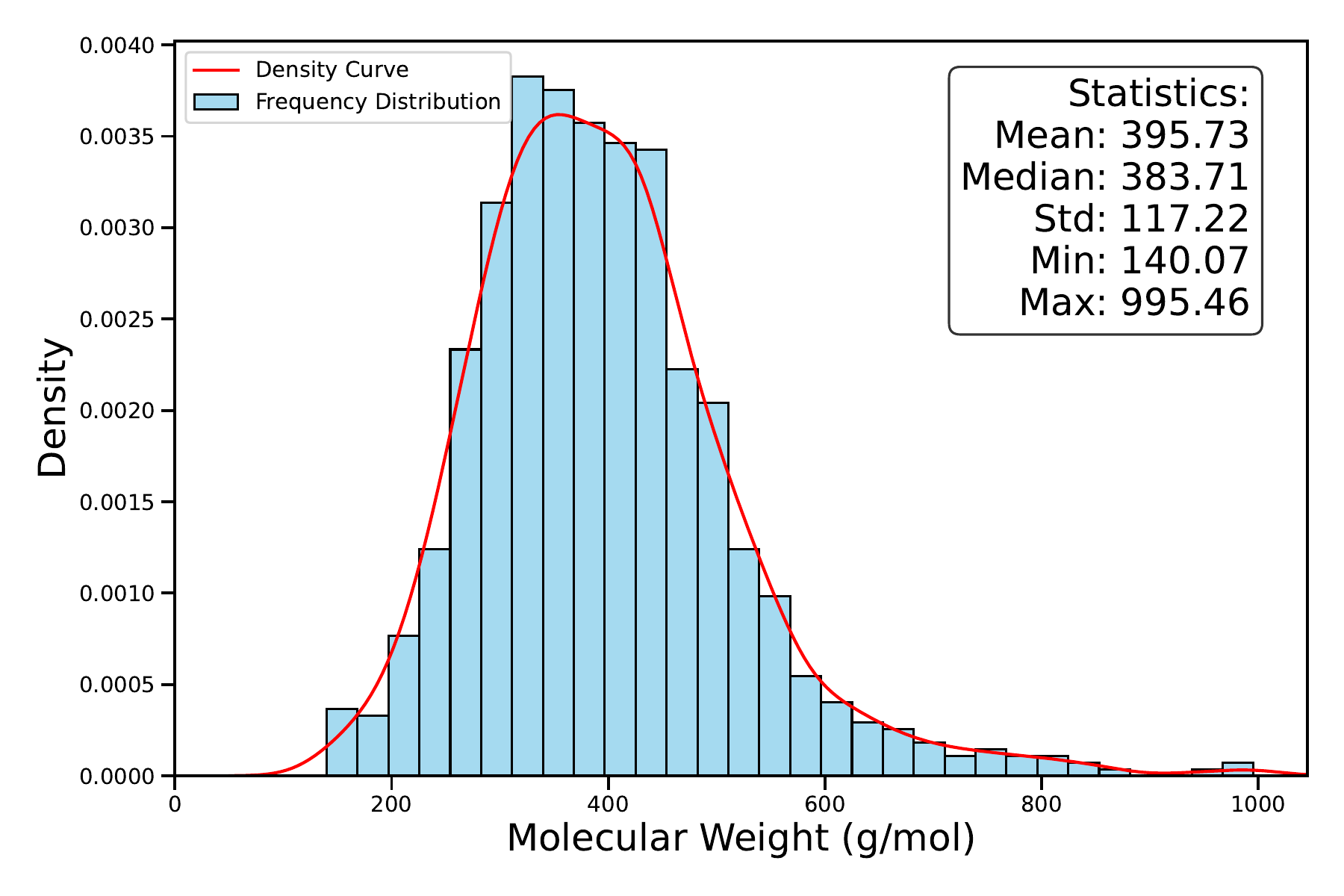}
    \caption{Distribution of molecular weights after recombination. The histogram reveals that the molecular weights of recombined products primarily fall within the 320–480 g/mol range, accounting for 54.89\% of the molecules. 79.94\% of the molecules lie within the 250–500 g/mol range, which is considered optimal for drug-likeness.}
    \label{fig:weight_dist}
\end{subfigure}
\hfill
\begin{subfigure}[t]{0.45\textwidth}
    \centering
    \includegraphics[width=\linewidth]{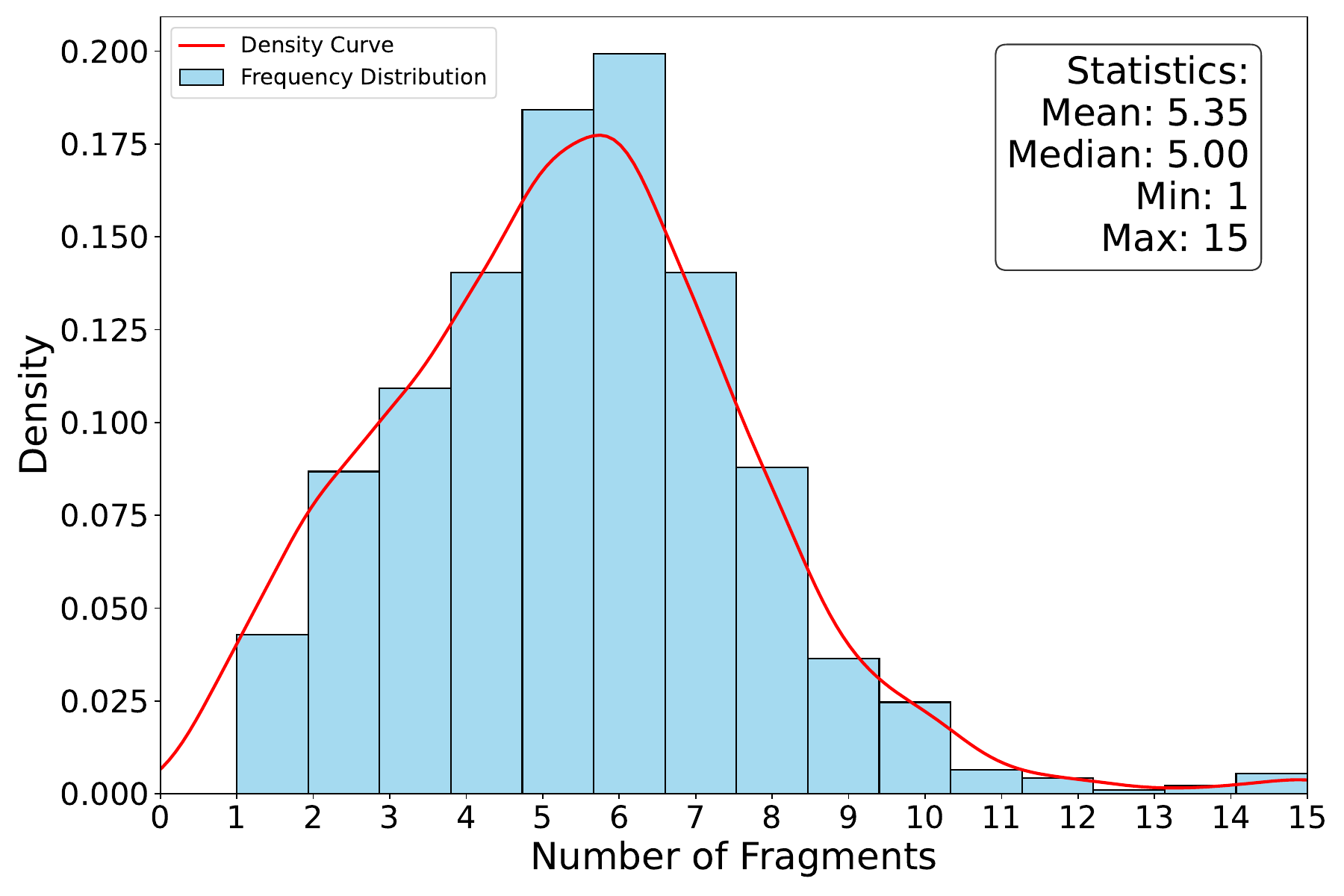}
    \caption{Distribution of molecular fragment numbers. The histogram shows that most molecules were fragmented into 1–10 parts, with the highest frequencies observed between 5 and 7 fragments. Fragment counts above 10 are rare but present, with less than 2\% of molecules exceeding this number.}
    \label{fig:fragment_dist}
\end{subfigure}
\caption{Analysis of molecular properties.}
\label{fig:combined_analysis}
\end{figure*}

\noindent\textbf{Result on fragment and recombination.} 
The results are presented in Figures \ref{fig:weight_dist} and \ref{fig:fragment_dist}. In the recombination task, the model generated 996 valid molecules with most molecular weights (54.89\%) falling in the 320–480 g/mol range. Notably, 79.94\% of the molecules were within the 250–500 g/mol range, considered drug-like and suggesting good drug-likeness in generated compounds. 
This reflects the model's effectiveness in producing compounds with high drug potential. 
For the fragment task, the model primarily generated fragment counts between 1 and 10, with fragmentation over 10 parts occurring in fewer than 2\% of cases.
This indicates the model’s capacity to fragment molecules into manageable parts while also handling larger molecules when needed, demonstrating versatility in fragmenting diverse molecular sizes.

\section{Case study on molecule recombination}
\label{app:C}
\begin{figure}[h]
    \centering
    \includegraphics[width=1\linewidth]{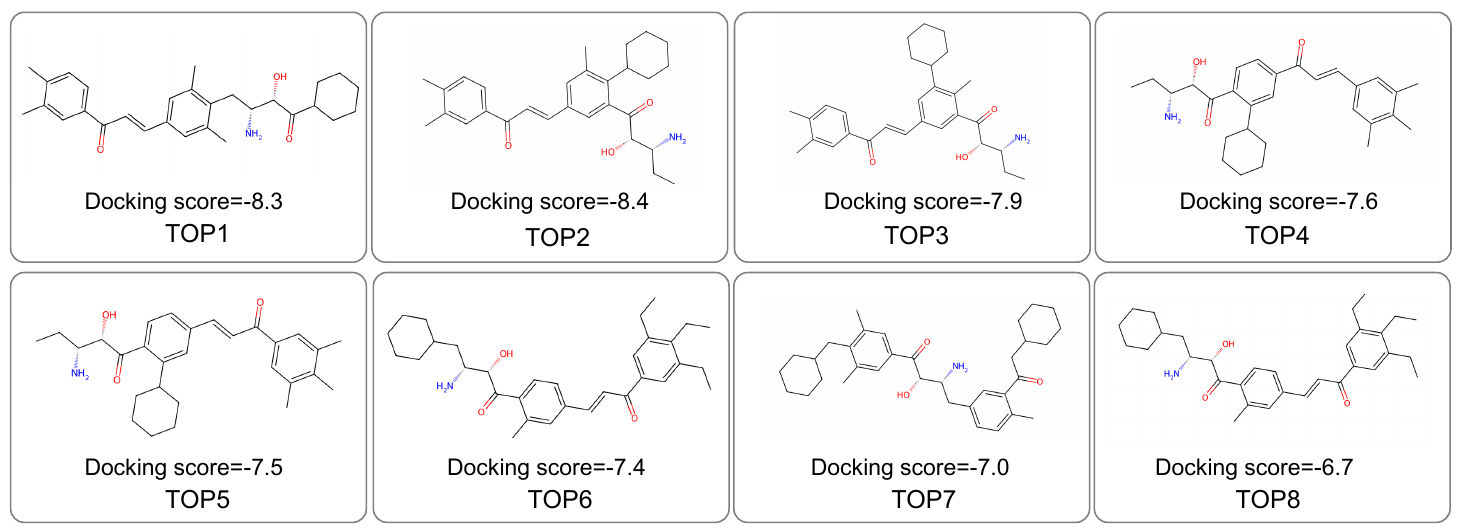}
    \caption{Different molecules generated by ChemDual. Docking score represent the maximum binding affinities (kcal/mol) between the ligands and MAP2 obtained through molecular docking.}
    \label{fig:case1}
\end{figure}

The molecules generated by ChemDual are shown in Figure \ref{fig:case1}, arranged in the order of their generation as determined by the docking score.
The generated compounds exhibit a high degree of similarity to original molecule in terms of molecular fingerprints, reflecting the effective inheritance of key structural and chemical features from the original fragments.

% \begin{figure*}[bt]
% \centering
% % 第一行 4 个子图
% \subfloat[-8.3]{\includegraphics[width=0.2\textwidth]{image/case1/molecule1}%
% \label{fig_case_1}}
% \hfil
% \subfloat[-7.9]{\includegraphics[width=0.2\textwidth]{image/case1/molecule2}%
% \label{fig_case_2}}
% \hfil
% \subfloat[-7.0]{\includegraphics[width=0.2\textwidth]{image/case1/molecule3}%
% \label{fig_case_3}}
% \hfil
% \subfloat[-7.5]{\includegraphics[width=0.2\textwidth]{image/case1/molecule4}%
% \label{fig_case_4}}

% \vspace{0.1em} % 两行之间的间距

% % 第二行 4 个子图
% \subfloat[-7.6]{\includegraphics[width=0.2\textwidth]{image/case1/molecule5}%
% \label{fig_case_5}}
% \hfil
% \subfloat[-8.4]{\includegraphics[width=0.2\textwidth]{image/case1/molecule6}%
% \label{fig_case_6}}
% \hfil
% \subfloat[-6.7]{\includegraphics[width=0.2\textwidth]{image/case1/molecule7}%
% \label{fig_case_7}}
% \hfil
% \subfloat[-7.4]{\includegraphics[width=0.2\textwidth]{image/case1/molecule8}%
% \label{fig_case_8}}

% \caption{Different molecules generated by ChemDual. The values below each structure (a-h) represent the maximum binding affinities (kcal/mol) between the ligands and Methionine aminopeptidase 2 obtained through molecular docking.}
% \label{fig:case1}
% \end{figure*}

\end{document}